\documentclass[10pt,twocolumn,letterpaper]{article}

\usepackage{cvpr}              

\usepackage{times}
\usepackage{helvet}
\usepackage{courier}
\usepackage{graphicx}
\usepackage{natbib}
\usepackage{caption}
\usepackage{algorithm}
\usepackage{algorithmic}

\usepackage{booktabs}       
\usepackage{amsfonts}       
\usepackage{nicefrac}       
\usepackage{xcolor}         

\usepackage{amsmath}
\usepackage{amssymb}
\usepackage{mathtools}
\usepackage{amsthm}

\usepackage{color}
\usepackage{bm}
\usepackage{makecell}
\usepackage{multirow}
\usepackage{adjustbox}
\usepackage{epstopdf}
\usepackage{threeparttable}
\usepackage{tablefootnote}
\usepackage{soul}

\usepackage{comment}

\DeclareMathAlphabet{\mathcal}{OMS}{cmsy}{m}{n}

\definecolor{cvprblue}{rgb}{0.21,0.49,0.74}
\usepackage[pagebackref,breaklinks,colorlinks,allcolors=cvprblue]{hyperref}
\usepackage[accsupp]{axessibility}

\def\paperID{} 
\def\confName{CVPR}
\def\confYear{2026}

\title{EvoComp: Learning Visual Token Compression for Multimodal Large \\ Language Models via Semantic-Guided Evolutionary Labeling}

\author{
    Jiafei Song\thanks{Equal Contribution} , 
    Fengwei Zhou\footnotemark[1] , 
    Jin Qu, 
    Wenjin Jason Li, 
    Tong Wu, 
    Gengjian Xue, \\
    Zhikang Zhao, 
    Daomin Wei, 
    Yichao Lu, 
    Bailin Na\thanks{Corresponding Author} \\
    OPPO CTG \\
    {\tt\small \{songjiafei, zhoufengwei, qujin, liwenjin1, wutong1, xuegengjian, } \\
    {\tt\small zhaozhikang, weidaomin1, yichao.lu, nabailin\}@oppo.com}
}

\begin{document}
\maketitle

\begin{abstract}
Recent Multimodal Large Language Models (MLLMs) have demonstrated strong performance on vision-language understanding tasks, yet their inference efficiency is often hampered by the large number of visual tokens, particularly in high-resolution or multi-image scenarios. To address this issue, we propose EvoComp, a visual token compression framework that significantly reduces token count while preserving task accuracy. EvoComp introduces a lightweight encoder-only transformer-based compressor that selects the most informative and non-redundant visual tokens by jointly considering visual and textual contexts. A core challenge lies in providing effective supervision for training the compressor. To this end, we design an evolutionary labeling strategy that searches for token subsets minimizing the MLLM's output loss, while enforcing semantic diversity through vocabulary-based token grouping. We further train the compressor using a tailored loss function combining the GHM loss to mitigate class and difficulty imbalance, and a cosine similarity regularization to encourage semantic separation between retained and discarded tokens. Extensive experiments across multiple vision-language benchmarks show that EvoComp outperforms existing methods based on attention or similarity heuristics. Notably, it retains 99.3\% of the original accuracy under 3x token compression and delivers up to 1.6x speedup on mobile devices.
\end{abstract}

\section{Introduction}

Multimodal Large Language Models (MLLMs)~\citep{alayrac2022flamingo,li2023blip,driess2023palm,liu2023visual,zhuminigpt} have progressed impressively in vision-language understanding tasks. They demonstrate strong capabilities in high-resolution image interpretation~\citep{liu2024improved,liu2024llavanext,wang2024qwen2,dong2024internlm}, multi-image reasoning~\citep{ye2024mplug,jiang2024mantis,wang2024longllava,wahed2024prima}, and even video-based comprehension~\citep{li2024llava,li2025llavaonevision,chen2024longvila,chen2024expanding}. 
However, these advances come with substantial computational burdens, as such tasks introduce a large number of visual tokens, often hundreds or thousands per input~\citep{liu2024improved,liu2024llavanext,li2024llava}. As a result, the inference cost of MLLMs scales dramatically in terms of memory consumption and latency, especially due to the quadratic complexity of transformer-based attention mechanisms~\citep{jin2024efficient,zhang2024treat}. This bottleneck becomes particularly severe in resource-constrained environments such as edge devices or mobile platforms~\citep{yao2024minicpm}. Therefore, efficient inference for MLLMs remains a critical challenge for real-world deployment.

\begin{figure*}[t]
  \centering
  \includegraphics[width=0.98\textwidth]{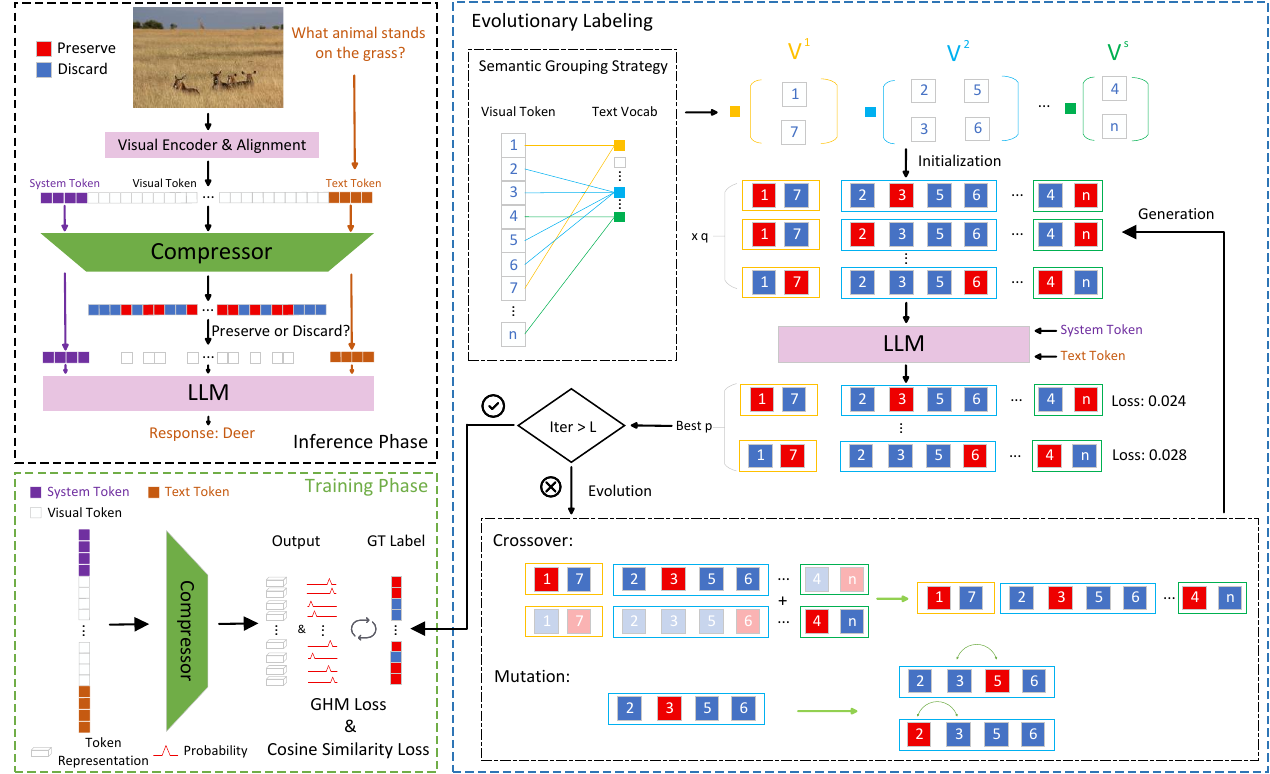}
  \caption{An overview of the EvoComp framework. \textbf{Evolutionary Labeling (Right)} searches for informative visual tokens that minimize the MLLM task loss, while ensuring non-redundancy by the semantic grouping strategy. \textbf{Training Phase (Left Bottom)} trains a lightweight compressor using the searched labels, optimized with a combination of GHM and cosine similarity loss. \textbf{Inference Phase (Left Top)} applies the trained compressor to filter tokens for efficient and accurate multimodal inference.}
  \label{fig:overview}
  \vspace{-4pt}
\end{figure*}

A key observation is that visual representations are often redundant, where a few tokens capture most of the semantic content~\citep{naseer2021intriguing,he2022masked,liang2022not}. This motivates visual token compression, that is, selecting or synthesizing a compact subset of tokens that preserves the model performance while reducing the inference cost~\citep{bolya2022tome,chen2024image,liu2024multi}. 

Recent methods for visual token compression typically rely on attention scores to identify important tokens~\citep{chen2024image,arif2025hired,Liu2025:GlobalCom2} or similarity metrics to remove redundant tokens~\citep{bolya2022tome,jiang2025kind,wen2025stop}.
Attention-based approaches suffer from position bias~\citep{liu2024lost,gu2025when,wu2025on,wen2025token}, and are often incompatible with efficient attention operators like FlashAttention~\citep{Yang_2025_CVPR,wen2025token}, while similarity-based filtering ignores the semantic importance of tokens~\citep{wen2025token}. 
A third category incorporates text-conditioned token selection~\citep{zhang2024sparsevlm,xing2024pyramiddrop,liu2024multi} via cross-modal attention, yet still depends on attention-based heuristics rather than directly optimizing for the model's output correctness.

In this work, we propose EvoComp, a framework for visual token \textbf{Comp}ression that learns to select informative and diverse tokens via \textbf{Evo}lutionary labeling. An overview of our method is illustrated in Figure~\ref{fig:overview}. Given an MLLM consisting of a visual encoder, an LLM and an alignment module between them~\citep{liu2023visual,liu2024improved,liu2024llavanext}, we insert a trained transformer-based compressor between the alignment module and the LLM. It receives aligned embeddings, both visual and textual, and then outputs retention probabilities to guide visual token selection.

The central challenge lies in constructing high-quality supervision signals to train the compressor. Instead of using heuristics, we generate supervision via an evolutionary algorithm~\citep{goldberg1989genetic,petrowski2017evolutionary,real2017large} applied to existing MLLM training data~\citep{liu2024improved}. Specifically, we search for a binary mask over visual tokens, i.e. a 0-1 vector indicating which tokens to keep or drop, such that the task-specific loss of the MLLM is minimized with the retained visual tokens and all text tokens. In this way, the token selection is directly aligned with the task objective and is conditioned on the textual input. 
To further promote diversity, we propose a semantic grouping strategy that groups visual tokens based on their semantic similarity and constrains the search to permit only one token per group. 
The evolutionary algorithm offers exceptional flexibility to support various constraints, tasks, and MLLM architectures. 

The resultant binary masks are used to train the compressor. Given that retained tokens are a small minority of the original set, we address class and difficulty imbalance during training by using the GHM loss~\citep{li2019gradient}, which dynamically reweighs the gradients based on token difficulty. We also adopt a cosine similarity loss~\citep{zhou2025mooscomp} to penalize similarity between the retained and pruned token representations, thus improving the classification performance. Once trained, the compressor enables efficient token selection in a single forward pass.

In summary, our main contributions are as follows:
\begin{itemize}
\item We propose EvoComp for visual token compression, which constructs supervision signals for training a lightweight compressor to retain non-redundant and task-relevant tokens based on both visual and textual inputs.
\item We introduce an evolutionary algorithm to search for high-quality binary masks that align token retention with task-specific loss of the MLLM, integrating both task relevance and redundancy elimination.
\item We tailor a loss function that combines the GHM and cosine similarity losses to improve compressor training challenged by difficulty imbalance and token similarity.
\item Experiments on six vision-language understanding benchmarks demonstrate that under 3x to 9x visual token compression, EvoComp preserves 99.3\% to 94.9\% of the baseline accuracy before compression, while achieving 1.6x to 2.0x inference speedup on a smartphone.
\end{itemize}

\section{Related Work}

\noindent\textbf{Vision-Information-Based Visual Token Compression.} A growing body of research has explored visual token compression in MLLMs by analyzing token importance and diversity. 
A popular class of methods estimates token importance with attention scores. 
FastV~\citep{chen2024image} prunes visual tokens with lower attention scores after early layers of an MLLM.
HiRED~\citep{arif2025hired} uses class-to-patch attention to retain top-ranked tokens.
$\text{GlobalCom}^2$~\citep{Liu2025:GlobalCom2} also adopts class token attention and guides token compression using thumbnails.
However, attention scores suffer from position bias~\citep{liu2024lost,gu2025when,wu2025on,wen2025token} and may not accurately reflect semantic importance.

Another category of methods focuses on redundancy reduction through similarity-based filtering. 
ToMe~\citep{bolya2022tome} merges tokens with highly similar embeddings to minimize redundancy. 
G-Prune~\citep{jiang2025kind} constructs a graph to represent token similarities and performs information propagation to identify the most representative tokens. 
DART~\citep{wen2025stop} retains tokens dissimilar to the selected pivot tokens. 
However, such methods may retain diverse but unimportant tokens, as similarity alone does not guarantee significance~\citep{wen2025token}.

To take into account both importance and diversity, PruMerge \citep{shang2024llava} combines attention-based and similarity-based scoring. It first selects important visual tokens based on class token attention and then enhances retained tokens via key-based similarity clustering. 
Similarly, VisionZip~\citep{yang2025visionzip} selects the dominant tokens based on their attention scores and merges the remaining semantically similar tokens. 
In contrast to these methods, our method leverages an evolutionary algorithm to construct supervision labels that reflect both token importance and non-redundancy via task loss minimization and similarity-constrained search, respectively.

\noindent\textbf{Text-Conditioned Visual Token Compression.} Another line of work focuses on incorporating textual context into the process of visual token selection to retain tokens that are most relevant to the language prompt. 
By selecting vision-relevant text tokens and computing their aggregated attention to each visual token, SparseVLM~\citep{zhang2024sparsevlm} prunes the least relevant tokens layer by layer. 
PyramidDrop~\citep{xing2024pyramiddrop} retains all visual tokens in shallow layers and gradually drops less important ones in deeper layers using their attention scores to the last-instruction token.

In addition, there are works jointly considering token saliency and prompt alignment. For example, MustDrop~\citep{liu2024multi} identifies vision-critical tokens and merges similar neighboring tokens during visual encoding. In the prefilling stage, visual tokens semantically unrelated to the prompt are pruned. 
Although these methods adapt token selection to the prompt, they still largely depend on attention-based heuristics.
In contrast, EvoComp unifies the benefits of text-aware guidance, redundancy reduction, and output-oriented optimization.

\section{Methodology}

\subsection{Preliminary}\label{sec:preliminary}

MLLMs integrate visual and textual modalities to enable sophisticated vision-language understanding and reasoning. An MLLM typically consists of three components: a visual encoder, a modality alignment module, and an LLM~\citep{liu2023visual,liu2024improved,liu2024llavanext}.
The visual encoder (e.g., CLIP~\citep{radford2021learning} or SigLIP~\citep{zhai2023sigmoid}) transforms a visual input into a sequence of visual tokens, each representing a local patch. 
The alignment module (often a projection layer or a lightweight adapter) maps these tokens into a shared latent space compatible with the LLM input space to generate aligned visual tokens. 
Textual input includes system prompts, user queries, and, during training, optional gold-standard responses as supervision. These are embedded and concatenated with the aligned visual tokens before being fed into the LLM for generation or prediction.
As visual resolution increases or when multiple images are involved~\citep{liu2024improved,liu2024llavanext,li2024llava}, visual tokens often significantly outnumber text tokens, leading to poor inference efficiency~\citep{jin2024efficient,zhang2024treat}.

To address this issue, our work focuses on visual token compression, with the aim of reducing the number of visual tokens to accelerate inference while preserving task performance. 
We introduce a lightweight compressor module, placed between the alignment module and the LLM. It takes as input both aligned visual and text tokens, and outputs retention probabilities to select informative and non-redundant visual tokens. This design ensures full compatibility with existing MLLM pipelines without any modification or fine-tuning of the visual encoder, the alignment module, or the LLM itself, and enables plug-and-play deployment.

\subsection{Label Construction for Visual Token Compression}\label{sec:evolution}

Training a high-quality visual token compressor requires reliable supervision. However, standard vision-language datasets lack token-level importance labels. 
In this regard, we propose a search-based labeling strategy using an evolutionary algorithm~\citep{goldberg1989genetic,petrowski2017evolutionary,real2017large} to generate token retention labels that are directly aligned with the MLLM's output loss. It also provides great flexibility, allowing various constraints on the supervision and adapting to different datasets and models without the need for backpropagation.

Given a pretrained MLLM, the supervision signals for compressor training can be constructed using either its instruction-tuning datasets~\citep{liu2024improved,liu2024llavanext} or task-specific supervision datasets reflecting realistic use cases. 
For each sample, 
let $\bm{V}=\{\bm{v}_i\}_{i=1}^n$ and $\bm{T}=\{\bm{t}_i\}_{i=1}^m$ denote the aligned visual and text token embeddings (including both the input and the gold-standard response), respectively. 
The objective is to find a binary mask $\bm{m} \in \{0,1\}^n$ that indicates which visual tokens to retain, so that the resulting task-specific loss $\mathcal{L}(\bm{m})$ computed on the response tokens is minimized when the LLM receives the selected visual tokens and the full input text tokens. 
This aligns token selection directly with output quality, rather than relying on proxy metrics alone, such as attention or similarity.

To avoid selecting semantically redundant tokens, a constraint is imposed on the search space. We introduce a semantic grouping strategy that groups visual tokens based on semantic similarity, and enforce one selection per group. This encourages diversity and significantly reduces the size of the search space, as the evolutionary process does not need to make independent decisions for every token.

\begin{figure}[t]
  \centering
  \includegraphics[width=0.8\columnwidth]{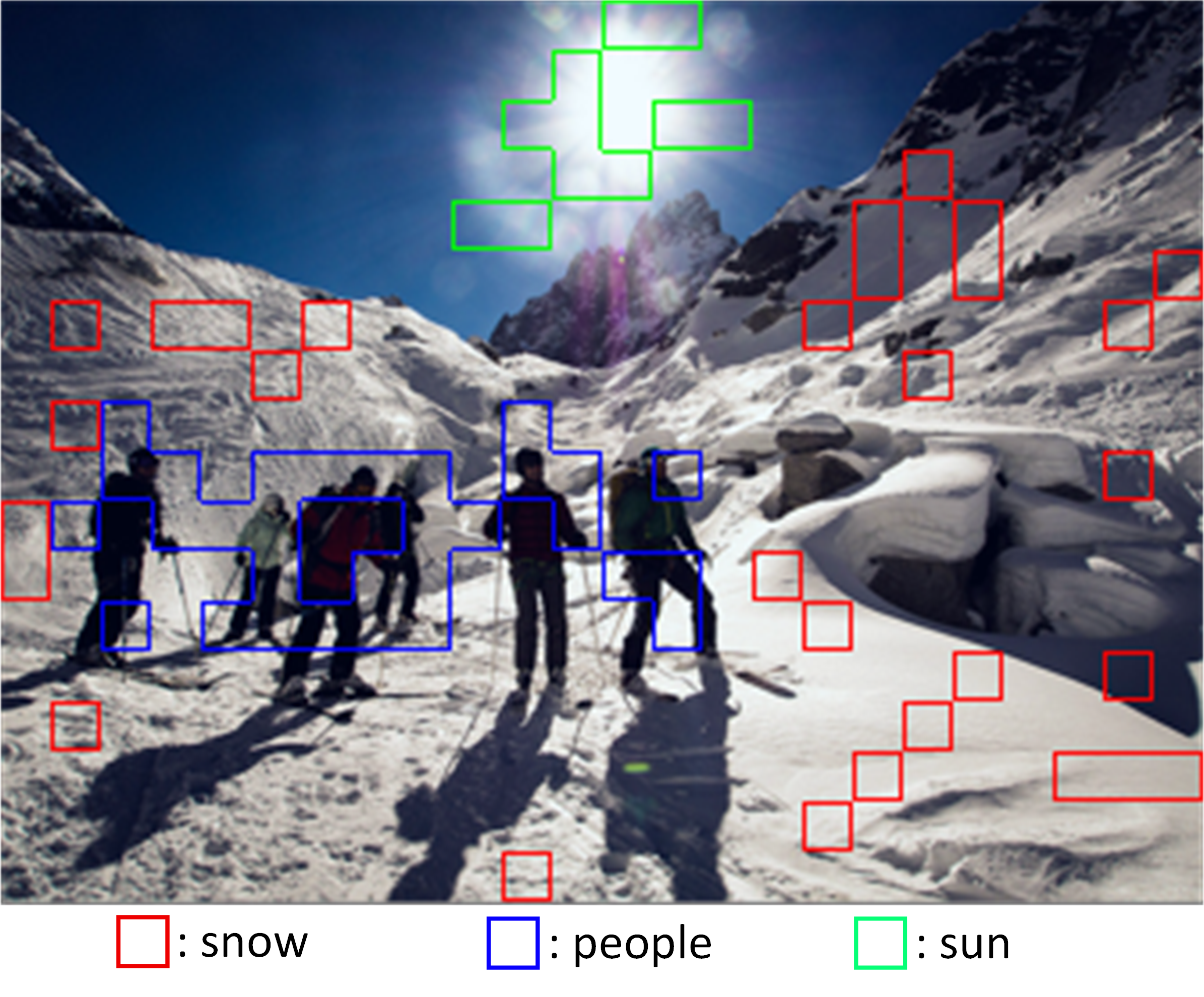}
  \caption{An example of semantic grouping result. Three representative subsets of visual tokens and their corresponding nearest vocabulary tokens are illustrated.}
  \label{fig:semantic_grouping}
  \vspace{-10pt}
\end{figure}

For clustering of visual tokens, LLM vocabulary embeddings $\bm{E}=\{\bm{e}_i\}_{i=1}^c$ can serve as semantically meaningful anchors, since visual tokens typically cluster around relevant embeddings in $\bm{E}$ and those that are closest to the same $\bm{e}_i$ are semantically similar~\citep{peng2024multi,liao2025langbridge}. Hence, we propose to measure the cosine similarity between the visual tokens and the vocabulary embeddings and group the visual tokens that share the same nearest $\bm{e}_i$ into subsets. Specifically, let $S_{ij} = \frac{\bm{v}_i \cdot \bm{e}_j}{\|\bm{v}_i\|_2 \|\bm{e}_j\|_2}$ denote the cosine similarity between $\bm{v}_i$ and $\bm{e}_j$. Then, two visual tokens $\bm{v}_i$ and $\bm{v}_k$ are grouped into one subset if and only if 

\noindent
\scalebox{0.9}{\parbox{1.11\columnwidth}{
\begin{equation}\label{eq:similar}
  \arg\max_j S_{ij} = \arg\max_j S_{kj}.
\end{equation}}}
According to this principle, the set $\bm{V}$ can be partitioned into disjoint subsets 

\noindent
\scalebox{0.9}{\parbox{1.11\columnwidth}{
\begin{equation}\label{eq:subset}
  \bm{V}^j=\{\bm{v}_{i_1^j}, \bm{v}_{i_2^j}, \cdots, \bm{v}_{i_{n_j}^j}\}, \, j=1, 2, \cdots, s,
\end{equation}}}
where $n_j$ is the size of $\bm{V}^j$, $i_k^j$ ($k=1,2,\cdots,n_j$) is the index of $\bm{v}_{i_k^j}$ (the $k$-th visual token of $\bm{V}^j$) in the original $\bm{V}$, and $s \leq c$ is the number of subsets. 
An example of a semantic grouping result is shown in Figure~\ref{fig:semantic_grouping}.
For each $\bm{V}^j$, we search for a binary mask $\bm{m}^j \in \{0,1\}^{n_j}$ with a single 1 in it, indicating which token to retain. These sub-masks are combined to form a global mask $\bm{m} = [\bm{m}^1, \bm{m}^2, \cdots, \bm{m}^s]$ over all visual tokens.

Within this structured search space, we first randomly initialize a population of candidate masks $\bm{M}=\{\bm{m}_i\}_{i=1}^q$ with population size $q$, and evaluate these candidates by computing the losses $\{\mathcal{L}(\bm{m}_i)\}_{i=1}^q$ of the LLM. 
This evaluation step can be efficiently parallelized, as each mask corresponds to an independent inference call. Based on $\{\mathcal{L}(\bm{m}_i)\}_{i=1}^q$, we select the top-$p$ candidates with the lowest losses to serve as parents $\bm{P}=\{\bm{m}_{i_k}= [\bm{m}^1_{i_k}, \bm{m}^2_{i_k}, \cdots, \bm{m}^s_{i_k}]\}_{k=1}^p$ for the next generation. The new candidate masks are then generated through two standard genetic operators:

\begin{itemize}
\item Crossover: Given a mask $\bm{m}_{i_k} = [\bm{m}^1_{i_k}, \bm{m}^2_{i_k}, \cdots, \bm{m}^s_{i_k}]$, crossover is performed with probability 0.9. If crossover is triggered, another mask $\bm{m}_{i_j} = [\bm{m}^1_{i_j}, \bm{m}^2_{i_j}, \cdots, \bm{m}^s_{i_j}]$ is sampled from the parents, and a new child mask $\bm{m}' = [\bm{m}^1_{i_k}, \bm{m}^2_{i_k}, \cdots, \bm{m}^{\lfloor s/2 \rfloor}_{i_k}, \bm{m}^{\lfloor s/2 \rfloor+1}_{i_j}, \cdots, \bm{m}^s_{i_j}]$ is generated by taking the first half of sub-masks from $\bm{m}_{i_k}$ and the second half from $\bm{m}_{i_j}$.

\item Mutation: Given a mask $\bm{m}_{i_k} = [\bm{m}^1_{i_k}, \bm{m}^2_{i_k}, \cdots, \bm{m}^s_{i_k}]$, mutation is performed with probability 0.2 for each sub-mask $\bm{m}^j_{i_k}$. If mutation is triggered, the 1 in $\bm{m}^j_{i_k}$ is randomly shifted to the left or right by one position.
\end{itemize}
After obtaining $q$ new candidate masks through crossover and mutation, we evaluate their losses using the same parallel inference strategy as before. Thus, high-loss parents are replaced with improved candidates.

At the end of $L$ iterations, the candidate with the lowest loss is selected as the supervision label for the current training sample. The evolutionary labeling algorithm is summarized in Algorithm~\ref{alg:evo_search}.
In our experiments, the population size $q$ is 48, the number of parents $p$ is 12, and the number of iterations $L$ is 10. Mask evaluation can be performed in batches. As samples are processed independently, the search scales well across GPUs, reducing the wall-clock time required over large-scale datasets.

\begin{algorithm}[t]\small
    \caption{Evolutionary Labeling for Visual Token Compression}
    \label{alg:evo_search}
    \begin{algorithmic}[1]
        \REQUIRE 
            Visual tokens $\bm{V}$; Text tokens $\bm{T}$; Vocabulary embeddings $\bm{E}$; Pretrained LLM; Population size $q$; Number of parents $p$; Number of iterations $L$;
        \ENSURE Binary mask $\bm{m}^*$ indicating selected visual tokens;
        \STATE Partition $\bm{V}$ into subsets $\{\bm{V}^j\}_{j=1}^s$ according to Eq.~\eqref{eq:similar};
        \STATE Randomly initialize a population of binary masks $\bm{M}^{(0)}=\{\bm{m}_i = [\bm{m}_i^1, \bm{m}_i^2, \cdots, \bm{m}_i^s]\}_{i=1}^q$, s.t., each $\bm{m}_i^j$ is one-hot;
        \STATE Evaluate task losses $\{\mathcal{L}(\bm{m}_i)\}_{i=1}^q$ for $\bm{M}^{(0)}$ in parallel;
        \STATE Select top-$p$ masks with lowest losses as parents $\bm{P}^{(0)}$;
        \FOR {$1 \leq l \leq L$}
            \STATE Initialize a new population $\bm{M}^{(l)} = \emptyset$;
            \WHILE {$|\bm{M}^{(l)}| < q$}
                \STATE Randomly sample a parent $\bm{m}_{\text{parent}}$ from $\bm{P}^{(l-1)}$;
                \STATE Generate $\bm{m}_{\text{crossover}}$ by performing crossover on $\bm{m}_{\text{parent}}$;
                \STATE Generate $\bm{m}_{\text{mutation}}$ by performing mutation on $\bm{m}_{\text{crossover}}$;
                \IF {$\bm{m}_{\text{mutation}}$ is not in $\bm{M}^{(0)}$}
                    \STATE Add $\bm{m}_{\text{mutation}}$ to $\bm{M}^{(0)}$;
                    \STATE Add $\bm{m}_{\text{mutation}}$ to $\bm{M}^{(l)}$;
                \ENDIF
            \ENDWHILE
            \STATE Evaluate task losses for $\bm{M}^{(l)}$ in parallel;
            \STATE Select top-$p$ masks from both $\bm{P}^{(l-1)}$ and $\bm{M}^{(l)}$ with lowest losses as parents $\bm{P}^{(l)}$;
        \ENDFOR
        \RETURN {$\bm{m}^* = \arg\min_{\bm{m} \in \bm{P}^{(L)}} \mathcal{L}(\bm{m})$}
    \end{algorithmic}
\end{algorithm}

\subsection{Supervised Learning for Visual Token Compression}\label{sec:learning}

With supervision signals obtained via evolutionary labeling, we can train the compressor to predict, for each visual token, whether it should be retained or not. A critical challenge is the presence of severe imbalances in both class distributions and difficulty levels, analogous to the foreground-background classification in object detection~\citep{lin2017focal,li2019gradient}. On the one hand, since visual representations contain significant redundancy, only a small number of visual tokens are labeled as ``positive'', i.e., should be retained, while the vast majority are redundant or background-like ``negative'' tokens. As a result, a large number of tokens are trivially easy to classify. However, there are also many tokens that are particularly hard to classify because of the high semantic variability of visual tokens. A visual token that is crucial in one image may be irrelevant in many others. Hence, uniformly treating all tokens can cause the model to bias toward learning from these easy and particularly hard tokens, thereby reducing training efficiency.

To address such a problem of attribute imbalance, we adopt the GHM loss~\citep{li2019gradient} for this token classification task. Specifically, let $f_{\bm{\varphi}}$ and $g_{\bm{\psi}}$ be the encoder-only transformer and the following classifier with parameters $\bm{\varphi}$ and $\bm{\psi}$, respectively. The transformer $f_{\bm{\varphi}}$ receives $\bm{V}$ and $\bm{T}$ (without the gold-standard response) and outputs visual and text token representations $\{\bm{h^v}_1, \cdots, \bm{h^v}_n, \bm{h^t}_1, \cdots, \bm{h^t}_m\} = f_{\bm{\varphi}}(\bm{V}, \bm{T})$. The encoder-only structure uses bidirectional attention to capture relationships between visual and text tokens. The classifier $g_{\bm{\psi}}$ is performed on the visual token representations to produce retention probabilities $\{p_i\}_{i=1}^n = g_{\bm{\psi}}(\{\bm{h^v}_i\}_{i=1}^n)$. The gradient norm $g_i$ is defined as $|p_i - y_i|$, where $y_i\in\{0, 1\}$ is the ground-truth label. 
The GHM loss dynamically adjusts each token's contribution based on its gradient density defined as

\noindent
\scalebox{0.9}{\parbox{1.11\columnwidth}{
\begin{equation}\label{eq:grad_density}
  GD(g_i) = \frac{1}{l_{\epsilon}(g_i)}\sum_{k=1}^n \delta_{\epsilon}(g_k, g_i),
\end{equation}}}
where $l_{\epsilon}(g_i) = \min(g_i+\epsilon/2, 1) - \max(g_i-\epsilon/2, 0)$ and $\delta_{\epsilon}(g_k, g_i)$ is an indicator function that is equal to 1 if $g_i-\epsilon/2 \leq g_k < g_i+\epsilon/2$ and to 0 otherwise. Intuitively, $GD(g_i)$ reflects the concentration of tokens whose gradient norm is close to $g_i$. Then the GHM loss is computed as 

\noindent
\scalebox{0.9}{\parbox{1.11\columnwidth}{
\begin{equation}\label{eq:GHMloss}
  \mathcal{L}_{\textbf{GHM-C}}(\bm{\varphi},\bm{\psi}) = \frac{1}{n} \sum_{i=1}^n \frac{n}{GD(g_i)}\ell(g_{\bm{\psi}}(\bm{h^v}_i), y_i),
\end{equation}}}
where $\ell(g_{\bm{\psi}}(\bm{h^v}_i), y_i)$ is the binary cross-entropy loss. The GHM loss is specifically designed to tackle class and difficulty imbalance problems by down-weighting the gradient contribution of easy negative and extremely hard samples.

To further assist the compressor in token classification, we encourage $f_{\bm{\varphi}}$ to learn discriminative representations of retained and pruned tokens, since $g_{\bm{\psi}}$ performs token classification based on these representations. To this end, we incorporate a cosine similarity loss that penalizes the similarity between the representations of the retained and discarded tokens. Specifically, let $\mathcal{I}_0=\{i \mid y_i=0\}$ and $\mathcal{I}_1=\{j \mid y_j=1\}$ be the indices of negative and positive tokens, respectively. Then the cosine similarity loss is computed as 

\noindent
\scalebox{0.9}{\parbox{1.11\columnwidth}{
\begin{equation}\label{eq:cosloss}
  \mathcal{L}_{\textbf{CS}}(\bm{\varphi}) = \frac{1}{|\mathcal{I}_0| |\mathcal{I}_1|} \sum_{i\in\mathcal{I}_0,j\in\mathcal{I}_1}\frac{\bm{h^v}_i \cdot \bm{h^v}_j}{\|\bm{h^v}_i\|_2 \|\bm{h^v}_j\|_2}.
\end{equation}}}
The adoption of this loss is closely tied to our label construction strategy, where each retained visual token is selected from a cluster of similar candidates while the remainders are pruned. As a result, these pruned tokens often share visual features similar to the retained one. However, the retained token contributes more to maintaining the accuracy of the MLLM output. Encouraging the compressor to distinguish informative tokens from semantically similar ones sharpens the classifier's decision boundary and improves overall selection quality~\cite{zhou2025mooscomp}.

The total loss function used to train the compressor is a weighted sum of the GHM and cosine similarity losses:

\noindent
\scalebox{0.9}{\parbox{1.11\columnwidth}{
\begin{equation}\label{eq:totalloss}
  \mathcal{L}(\bm{\varphi},\bm{\psi}) = \mathcal{L}_{\textbf{GHM-C}}(\bm{\varphi},\bm{\psi}) + \alpha \mathcal{L}_{\textbf{CS}}(\bm{\varphi}),
\end{equation}}}
where $\alpha$ balances the two terms. This combined objective improves the robustness and convergence behavior of the classifier and sharpens its decision boundary in dense and ambiguous regions of the embedding space.

Once trained, the compressor performs token selection in a single forward pass with negligible overhead. 
Given a target compression ratio, we select the top-$r$ visual tokens with the highest retention probabilities, where $r$ is determined by the ratio. 
Moreover, while token selection is completed before the tokens enter the LLM, the actual pruning can be flexibly applied at any layer within the LLM. This allows EvoComp to support early-layer pruning for maximum speedup or late-layer pruning for better performance, providing a high degree of flexibility during deployment.

\begin{table}[t]
    \centering
    \addtolength{\tabcolsep}{-0.4em}
    \caption{Performance of EvoComp and other methods on vision-language understanding with LLaVA-1.5-7B under different number of retained tokens.}
    \label{tab:llava-1.5-7b}
    \makebox[\columnwidth]{\resizebox{1.02\columnwidth}{!}{
    \begin{threeparttable}
    \begin{tabular}{lcccccc|c}
    \toprule
    \textbf{Method} & \textbf{GQA} & \textbf{MMB} & \textbf{MMB-CN} & \textbf{POPE} & $\textbf{VQA}^{\text{V2}}$ & \textbf{VizWiz} & \textbf{Avg.}  \\
    \hline
    \hline
    LLaVA-1.5-7B & \multicolumn{7}{c}{\textit{Upper Bound, 576 Tokens} \textbf{(100\%)}} \\
    \hline
    Vanilla & 61.9 & 64.7 & 58.1 & 85.9 & 78.5 & 50.0 & 100\%  \\
    \hline
    \hline
            & \multicolumn{7}{c}{\textit{Retain 192 Tokens} \textbf{($\downarrow$ 66.7\%)}} \\
    \hline
    FastV & 57.9 & \textbf{64.2} & \textbf{57.5} & 77.8 & 76.0 & 51.2 & 96.9\%  \\
    HiRED & 58.7 & 62.8 & 54.7 & 82.8 & 74.9 & 50.1 & 96.3\%  \\
    PruMerge & 54.3 & 59.6 & 52.9 & 71.3 & 70.6 & 50.1 & 90.7\%  \\
    VisionZip & 59.3 & 63.0 & 57.4 & \textbf{85.3} & 76.8 & 51.7 & 98.8\%  \\
    SparseVLM & 59.5 & 64.1 & 53.7 & \textbf{85.3} & 75.6 & 50.5 & 97.4\%  \\
    PyramidDrop & 57.3 & 63.3 & 56.8 & 82.3 & 75.1 & 51.1 & 97.0\%  \\
    MustDrop & 58.2 & 62.3 & 55.8 & 82.6 & 76.0 & 51.4 & 97.0\%  \\
    DART & 60.0 & 63.6 & 57.0 & 82.8 & 76.7 & 51.2 & 98.3\%  \\
    DART ($l=0$) & 57.1 & 59.9 & 53.6 & 78.4 & 73.2 & 51.7 & 94.2\%  \\
    \textbf{EvoComp} ($l=0$) & 59.7 & 63.4 & 56.1 & 85.0 & 76.6 & \textbf{52.3} & 98.7\%  \\
    \textbf{EvoComp} ($l=2$) & \textbf{60.8} & \textbf{64.2} & 57.2 & 84.1 & \textbf{77.4} & 51.7 & \textbf{99.3\%}  \\
    \hline
    \hline
            & \multicolumn{7}{c}{\textit{Retain 128 Tokens} \textbf{($\downarrow$ 77.8\%)}} \\
    \hline
    FastV & 55.9 & \textbf{63.7} & 56.6 & 72.0 & 73.3 & 51.4 & 94.4\%  \\
    HiRED & 57.2 & 61.5 & 53.6 & 79.8 & 73.4 & 51.3 & 94.8\%  \\
    PruMerge & 53.3 & 58.1 & 51.7 & 67.2 & 68.8 & 50.3 & 88.6\%  \\
    VisionZip & 57.6 & 62.0 & 56.9 & 83.2 & 75.6 & 52.1 & 97.4\%  \\
    SparseVLM & 58.4 & 64.5 & 51.1 & \textbf{85.0} & 73.8 & 51.4 & 96.3\%  \\
    PyramidDrop & 57.1 & 61.6 & 56.6 & 82.3 & 72.9 & 51.0 & 95.9\%  \\
    MustDrop & 56.9 & 61.1 & 55.2 & 78.7 & 74.6 & 52.1 & 95.4\%  \\
    DART & 58.7 & 63.2 & \textbf{57.5} & 80.1 & 75.9 & 51.7 & 97.5\%  \\
    DART ($l=0$) & 55.5 & 57.3 & 51.9 & 74.0 & 70.9 & 51.4 & 91.1\%  \\
    \textbf{EvoComp} ($l=0$) & 58.6 & 63.1 & 56.1 & 83.2 & 75.3 & \textbf{52.7} & 97.8\%  \\
    \textbf{EvoComp} ($l=2$) & \textbf{59.5} & 63.1 & 57.1 & 82.0 & \textbf{76.1} & 51.7 & \textbf{98.0\%}  \\
    \hline
    \hline
            & \multicolumn{7}{c}{\textit{Retain 64 Tokens} \textbf{($\downarrow$ 88.9\%)}} \\
    \hline
    FastV & 51.6 & 61.0 & 52.8 & 59.6 & 66.5 & 51.4 & 87.6\%  \\
    HiRED & 54.6 & 60.2 & 51.4 & 73.6 & 69.7 & 50.2 & 90.8\%  \\
    PruMerge & 51.9 & 55.3 & 49.1 & 65.3 & 67.4 & 50.1 & 86.0\%  \\
    VisionZip & 55.1 & 60.1 & \textbf{55.3} & 77.0 & 72.4 & \textbf{53.0} & 94.2\%  \\
    SparseVLM & 53.8 & 60.1 & 46.1 & 77.5 & 68.2 & 50.1 & 89.4\%  \\
    PyramidDrop & 47.5 & 58.8 & 50.5 & 55.9 & 69.2 & 50.7 & 84.9\%  \\
    MustDrop & 53.1 & 60.0 & 53.1 & 68.0 & 69.3 & 51.2 & 90.0\%  \\
    DART & 55.9 & 60.6 & 53.2 & 73.9 & 72.4 & 51.6 & 92.8\%  \\
    DART ($l=0$) & 52.5 & 52.6 & 45.4 & 65.7 & 66.9 & 50.5 & 84.5\%  \\
    \textbf{EvoComp} ($l=0$) & 56.1 & \textbf{61.9} & 55.1 & \textbf{78.2} & 72.0 & 52.7 & \textbf{94.9\%}  \\
    \textbf{EvoComp} ($l=2$) & \textbf{56.9} & \textbf{61.9} & 53.7 & 76.0 & \textbf{72.6} & 51.2 & 93.9\%  \\
    \bottomrule
    \end{tabular}
    \end{threeparttable}
    }}
    \vspace{-10pt}
\end{table}

\section{Experiments}

In this section, we evaluate the effectiveness and efficiency of EvoComp in visual token compression for MLLMs. We have conducted extensive experiments on various vision-language understanding benchmarks and target MLLMs. Our evaluations focus on assessing task accuracy under various compression ratios, measuring inference speedup on both GPU and edge NPU devices, and understanding the component contributions via ablation studies.

\begin{table}[t]
    \centering
    \addtolength{\tabcolsep}{-0.4em}
    \caption{Evaluation of extreme visual token compression for high-resolution images with LLaVA-NeXT-7B.}
    \label{tab:llava-next-7b}
    \makebox[\columnwidth]{\resizebox{1.02\columnwidth}{!}{
    \begin{threeparttable}
    \begin{tabular}{lcccccc|c}
    \toprule
    \textbf{Method} & \textbf{GQA} & \textbf{MMB} & \textbf{MMB-CN} & \textbf{POPE} & $\textbf{VQA}^{\text{V2}}$ & \textbf{VizWiz} & \textbf{Avg.}  \\
    \hline
    \hline
    LLaVA-NeXT-7B & \multicolumn{7}{c}{\textit{Upper Bound, 2880 Tokens} \textbf{(100\%)}} \\
    \hline
    Vanilla & 64.2 & 67.4 & 60.6 & 86.5 & 81.8 & 57.6 & 100\%  \\
    \hline
    \hline
            & \multicolumn{7}{c}{\textit{Retain 160 Tokens} \textbf{($\downarrow$ 94.4\%)}} \\
    \hline
    FastV & 53.5 & 60.3 & 53.1 & 62.5 & 67.9 & 50.8 & 84.0\%  \\
    VisionZip & 55.5 & 60.1 & 52.1 & 74.8 & 71.4 & \textbf{56.1} & 88.8\%  \\
    SparseVLM & 56.6 & 61.1 & 53.8 & \textbf{80.3} & 72.2 & 51.7 & 89.7\%  \\
    PyramidDrop & 53.0 & 60.0 & 53.4 & 66.4 & 66.6 & 51.2 & 84.5\%  \\
    MustDrop & 54.8 & 61.7 & 53.5 & 74.8 & 71.5 & 53.5 & 88.7\%  \\
    $\text{GlobalCom}^2$ & 54.1 & 59.1 & 50.9 & 75.8 & 70.0 & 50.2 & 86.1\%  \\
    DART & 57.3 & 61.2 & 52.6 & 77.2 & 73.6 & 52.8 & 89.6\%  \\
    DART ($l=0$) & 54.8 & 56.2  & 47.8 & 68.7 & 68.3 & 50.0 & 82.9\%  \\
    \textbf{EvoComp} ($l=0$) & \textbf{59.5} & \textbf{62.7} & \textbf{55.8} & 78.1 & \textbf{74.5} & 53.9 & \textbf{92.1\%}  \\
    \bottomrule
    \end{tabular}
    \end{threeparttable}
    }}
\end{table}

\begin{table}[tp]
    \centering
    \addtolength{\tabcolsep}{-0.4em}
    \caption{Evaluation of token selection transferability from LLaVA-1.5-7B to LLaVA-1.5-13B. The results of EvoComp are based on the transferred setting, whereas those of the other methods are derived from LLaVA-1.5-13B itself.}
    \label{tab:transferability}
    \makebox[\columnwidth]{\resizebox{1.02\columnwidth}{!}{
    \begin{threeparttable}
    \begin{tabular}{lcccccc|c}
    \toprule
    \textbf{Method} & \textbf{GQA} & \textbf{MMB} & \textbf{MMB-CN} & \textbf{POPE} & $\textbf{VQA}^{\text{V2}}$ & \textbf{VizWiz} & \textbf{Avg.}  \\
    \hline
    \hline
    LLaVA-1.5-13B & \multicolumn{7}{c}{\textit{Upper Bound, 576 Tokens} \textbf{(100\%)}} \\
    \hline
    Vanilla & 63.2 & 67.7 & 63.6 & 85.9 & 80.0 & 53.6 & 100\%  \\
    \hline
    \hline
            & \multicolumn{7}{c}{\textit{Retain 64 Tokens} \textbf{($\downarrow$ 88.9\%)}} \\
    \hline
    FastV & 55.8 & 64.7 & 60.1 & 69.4 & 71.2 & \textbf{55.1} & 91.8\%  \\
    VisionZip & 56.2 & 64.9 & \textbf{61.1} & 76.0 & 73.7 & 53.4 & 93.5\%  \\
    SparseVLM & 55.6 & 62.7 & 58.8 & 76.1 & 71.3 & 50.9 & 91.0\%  \\
    PyramidDrop & 53.5 & 60.8 & 53.0 & 69.8 & 69.0 & 46.9 & 85.5\%  \\
    MustDrop & 53.5 & 63.0 & 57.9 & 65.8 & 70.0 & 52.7 & 88.5\%  \\
    DART     & 57.1 & 65.5 & 61.0 & 75.6 & 73.6 & 54.6 & 94.1\%   \\
    DART ($l=0$) & 54.2 & 60.7 & 57.2 & 74.4 & 69.1 & 51.8 & 89.2\%  \\
    \textbf{EvoComp} ($l=0$) & 56.5 & \textbf{66.8} & 60.2 & 76.5 & 73.0 & 52.5 & 93.5\%  \\
    \textbf{EvoComp} ($l=2$) & \textbf{58.3} & 65.5 & 60.8 & \textbf{76.9} & \textbf{73.9} & 53.5 & \textbf{94.4}\%  \\
    \bottomrule
    \end{tabular}
    \end{threeparttable}
    }}
    \vspace{-10pt}
\end{table}

\subsection{Implementation Details}

\noindent\textbf{Target and Compression Models.} We evaluate EvoComp on the widely used MLLMs, including LLaVA-1.5-7B/13B~\citep{liu2024improved}, LLaVA-NeXT-7B~\citep{liu2024llavanext}, and Qwen2.5-VL-7B~\citep{bai2025qwen2}. The latter two are used to validate visual token compression for high-resolution images. 
The compressor consists of a single-layer transformer followed by a linear classifier. The transformer structure mirrors that of the LLM layer in the corresponding MLLM with two distinctions: 1) causal attention is replaced with bidirectional attention, and 2) a skip connection is incorporated to facilitate a direct link from the input to the output.

\noindent\textbf{Training and Evaluation Datasets.} For label construction and compressor training, we use a subset of the LLaVA-1.5 instruction-tuning mixture dataset~\citep{liu2024improved}.
For each target MLLM, we have constructed a labeled dataset to train their corresponding compressor. More training details can be found in Appendix~\ref{sec:training}.
For evaluation, we adopt six standard vision-language benchmarks, including GQA~\citep{hudson2019gqa}, MMBench/MMBench-CN (MMB/MMB-CN)~\citep{liu2024mmbench}, POPE~\citep{li2023evaluating}, VQAv2 ($\text{VQA}^{\text{V2}}$)~\citep{goyal2017making}, and VizWiz~\citep{gurari2018vizwiz}. Additional details can be found in Appendix~\ref{sec:dataset}.

\noindent\textbf{Baselines.} We compare EvoComp with several powerful visual token compression methods, including attention-based token importance estimation (FastV~\citep{chen2024image}, HiRED~\citep{arif2025hired}, $\text{GlobalCom}^2$~\citep{Liu2025:GlobalCom2}), similarity-based filtering (PruMerge~\citep{shang2024llava}, VisionZip~\citep{yang2025visionzip}, DART~\citep{wen2025stop}), and text-conditioned token selection (SparseVLM~\citep{zhang2024sparsevlm}, PyramidDrop~\citep{xing2024pyramiddrop}, MustDrop~\citep{liu2024multi}).

\begin{figure}[t]
	\begin{subfigure}{.23\textwidth}
		\centering
		\includegraphics[width=.98\linewidth]{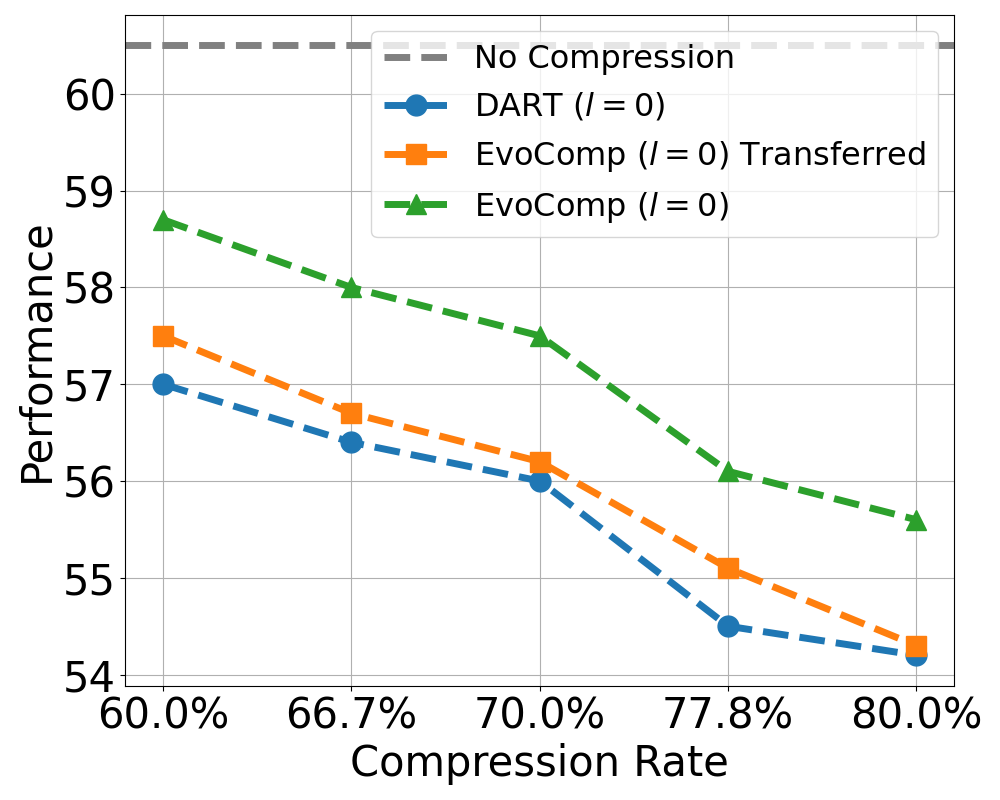}
		\caption{GQA}
		\label{fig:gqa}
	\end{subfigure}
	\begin{subfigure}{.23\textwidth}
		\centering
		\includegraphics[width=.98\linewidth]{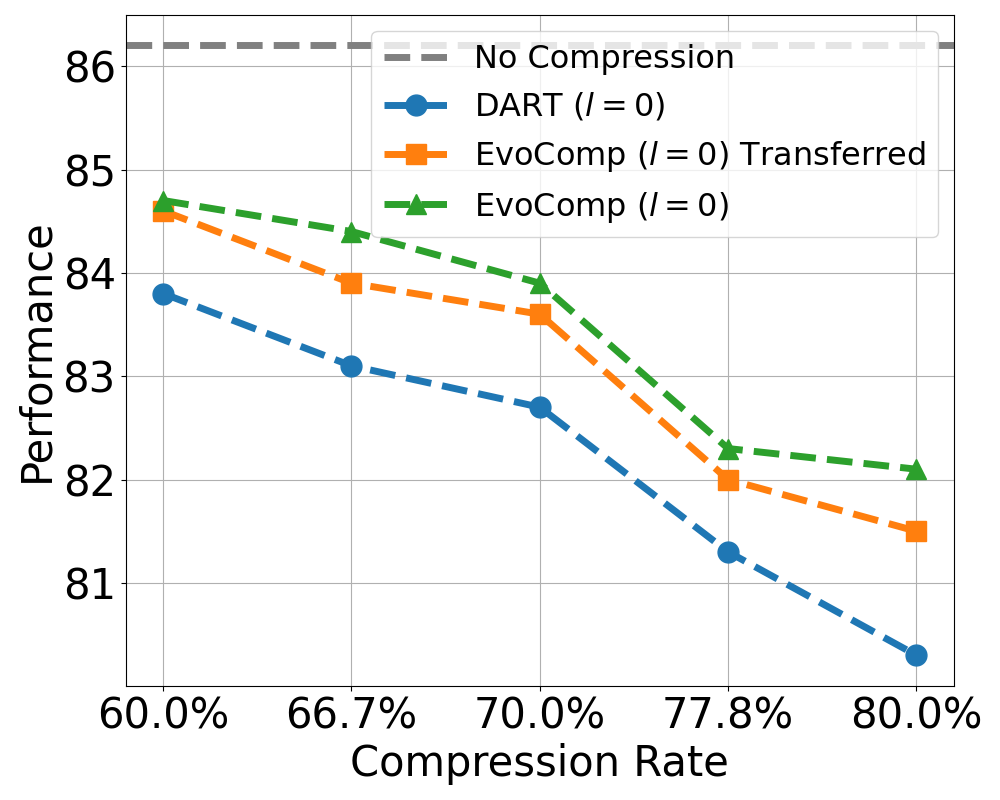}
		\caption{POPE}
		\label{fig:pope}
	\end{subfigure}
	\caption{Evaluation of compressor transferability from LLaVA-1.5-7B to Qwen2.5-VL-7B.}
	\label{fig:qwen}
	\vspace{-10pt}
\end{figure}

\subsection{Main Results}

\noindent\textbf{Evaluation of Supervised Learning for Visual Token Compression.} We evaluate EvoComp on various vision-language understanding tasks with LLaVA-1.5-7B and LLaVA-NeXT-7B. To illustrate the effectiveness and flexibility of EvoComp, we have evaluated two token compression strategies: before the LLM (EvoComp ($l=0$)) and after its second transformer layer (EvoComp ($l=2$)). As shown in Table~\ref{tab:llava-1.5-7b} and Table~\ref{tab:llava-next-7b}, EvoComp consistently outperforms other competitors on average with multiple token retention ratios. 
When dropping 88.9\% of the visual tokens, EvoComp maintains strong performance with only 5.1\% accuracy loss for LLaVA-1.5-7B. Even under an extreme 18x compression ratio, EvoComp maintains 92.1\% of the baseline accuracy for LLaVA-NeXT-7B. These results highlight the effectiveness of evolutionary labeling and supervised training in selecting informative and diverse tokens for accurate multimodal inference. The tokens selected by EvoComp are visualized in Appendix~\ref{sec:token_selection}.

Moreover, EvoComp ($l=0$) achieves a comparable or even better accuracy than EvoComp ($l=2$) and other methods that prune tokens within intermediate LLM layers. 
Also notable is that it outperforms DART with tokens dropped prior to LLM inference (DART ($l=0$)) by a large margin. 
By decoupling token compression from the LLM internals, EvoComp enables a plug-and-play compression module compatible with off-the-shelf LLMs and supports deployment flexibility across hardware platforms, where fine-grained control over LLM layers may not be feasible.

\noindent\textbf{Transferability of Visual Token Selection across MLLMs.} 
To reduce system complexity and enable scalable deployment across different model scales, we investigate whether visual tokens selected using a smaller MLLM can be reused by a larger one within the same family. Specifically, we leverage both the visual tokens produced by LLaVA-1.5-7B and the token selection decisions made by the compressor trained on LLaVA-1.5-7B and apply the same selected token indices to LLaVA-1.5-13B for inference. 
As illustrated in Table~\ref{tab:transferability}, this selection strategy remains highly effective, demonstrating that visual token importance and diversity evaluation can be largely decoupled from specific downstream MLLMs. This indicates that our compressor captures intrinsic visual semantics that generalize across model scales.
This also opens up opportunities where visual token selection can be performed by a lightweight MLLM with a learned compressor and reused by larger MLLMs.

\noindent\textbf{Transferability of Compressor across MLLMs.} To examine the transferability of the compressor between heterogeneous MLLMs, we apply the compressor trained on LLaVA-1.5-7B directly to Qwen2.5-VL-7B without any fine-tuning (EvoComp ($l=0$) Transferred). Since the visual and text token embeddings produced by the two models have different hidden dimensions, we apply adaptive average pooling to the input embeddings of the compressor, enabling it to operate on variable-sized feature vectors across different models. We also train a compressor for Qwen2.5-VL-7B model and test the results for comparison with the transferred outcomes.
As illustrated in Figure~\ref{fig:qwen}, the transferred compressor remains highly effective when applied to Qwen2.5-VL-7B. EvoComp achieves competitive performance across various compression ratios. Even without any retraining on the target model, EvoComp demonstrates competitive accuracy, underscoring its robustness and generality.
This suggests that the token selection learned by EvoComp captures semantic and task-relevant structures that generalize across different model configurations, making it practical for plug-and-play deployment.

\begin{figure}[t]
   \centering
   \includegraphics[width=0.7\columnwidth]{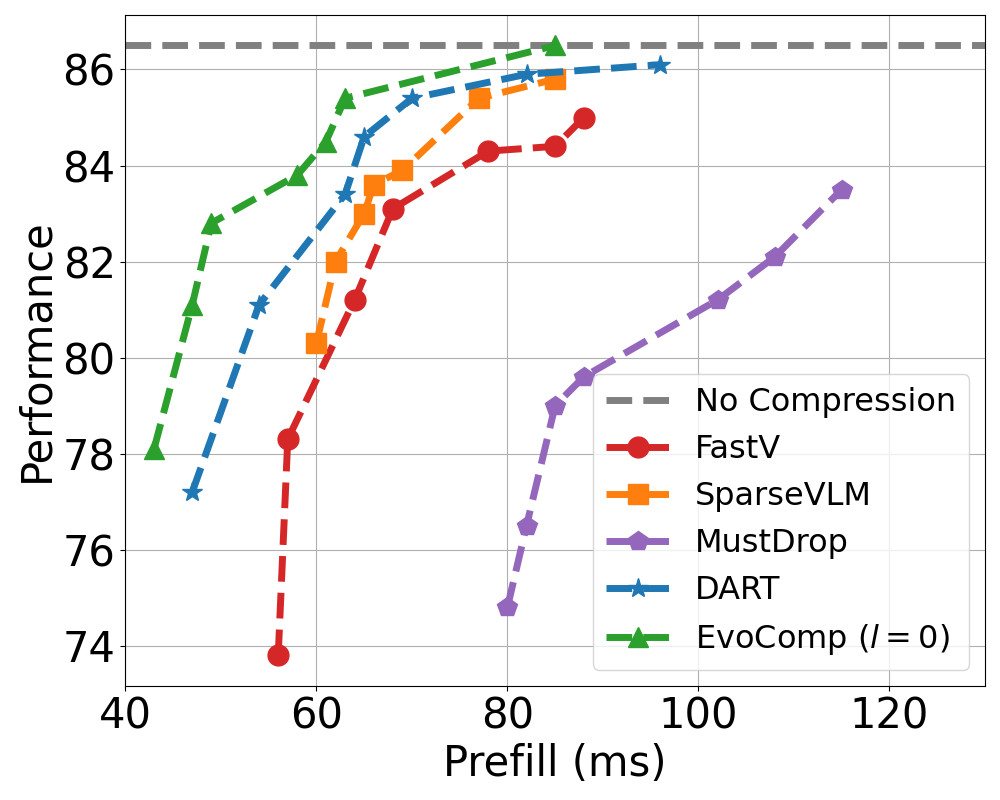}
   \caption{Speed-performance of LLaVA-NeXT-7B enabled by visual token compression on POPE. }
   \label{fig:pareto}
   \vspace{-15pt}
\end{figure}

\noindent\textbf{Inference Speedup on NVIDIA GPUs.} To assess the inference acceleration enabled by visual token compression, we benchmark LLaVA-NeXT-7B on POPE using an A100-80G GPU. Figure~\ref{fig:pareto} illustrates the trade-off between task accuracy and LLM prefill latency (including token selection) under different compression ratios. EvoComp achieves better accuracy than other methods at each latency level. Compared with the uncompressed setting, where the prefill latency and total runtime are 175 ms and 246 ms, respectively, EvoComp reduces them to 43 ms and 93 ms, achieving 4.1× speedup in prefill and 2.6× overall acceleration.
Our pruning strategy allows the LLM to run with a shorter sequence from the very first layer, without requiring any dynamic adjustment during inference.
These results highlight EvoComp's ability to combine high accuracy under aggressive compression with practical latency gains.

\noindent\textbf{Inference Speedup on Mobile Devices.} Accelerating large model inference is especially critical on resource-constrained edge devices. To further assess the practical benefits of visual token compression, we deploy LLaVA-1.5-7B and its compressor on a smartphone equipped with a dedicated NPU, following standard on-device optimization and compilation procedures (see Appendix~\ref{sec:llm_deployment} for details).
We use the GQA example and apply EvoComp to reduce visual tokens to fixed lengths of 192, 128, and 64. As shown in Table~\ref{tab:latency_npu}, compression yields up to 2.0x speedup in total latency and 2.2x acceleration in the combined time of visual encoding, token compression and LLM prefill, leading to a significant reduction of the Time-To-First-Token (TTFT).

\begin{table}[t]
    \centering
    \caption{Latency evaluation of the compressor and LLaVA-1.5-7B on a smartphone.}
    \label{tab:latency_npu}
    \resizebox{\columnwidth}{!}{
    \begin{tabular}{lcccc}
    \toprule
    \# Retained Visual Tokens  & 576 & 192 & 128 & 64    \\
    \midrule
    Visual Encoding (ms)    &  220  &  220  &   218   &  218     \\
    Compression (ms)        &  -    &  121  &   121   &  121     \\
    LLM Prefill (ms)        &  853  &  310  &   309   &  156  \\
    LLM Decoding (ms)       &  81   &  75   &   75    &  74  \\
    Total (ms)              &  1154 &  726  &   723   &  569  \\
    \midrule
    Speedup                 &  1.0x   &  1.6x &   1.6x  &  2.0x  \\
    \bottomrule
    \end{tabular}
    }
    \vspace{-5pt}
\end{table}

\subsection{Ablation Studies}

To better understand the contributions of different components in EvoComp, we perform a series of ablation studies on LLaVA-1.5-7B using MMB and MMB-CN. The results are reported in Table~\ref{tab:ablation_contribution}.

\noindent\textbf{Label Quality Analysis.} We first examine the benefit of our evolutionary labeling strategy by comparing EvoComp against 1) \textit{Random Labeling}: employing the same semantic grouping strategy while randomly selecting one token per group to be retained; 2) \textit{EvoComp w/ DPC-KNN}: clustering visual tokens utilizing DPC-KNN~\citep{du2016study}, a k-nearest-neighbor based density peaks clustering method, and retaining the cluster center tokens.
Our search-based labels consistently yield better supervised compressor performance than random labels as they are directly optimized for the final task loss. The labels constructed by DPC-KNN outperform random labels but still underperform compared to our evolutionary search strategy with semantic grouping. Moreover, unlike DPC-KNN, our grouping strategy avoids extra hyperparameters (e.g., number of neighbors and clusters) that are difficult to tune and can introduce tuning bias.

\begin{table}[t]
    \centering
    \caption{Ablation study on vision-language understanding using LLaVA-1.5-7B.}
    \label{tab:ablation_contribution}
    \resizebox{\columnwidth}{!}{
    \begin{tabular}{lcccc}
    \toprule
    Method & \multicolumn{2}{c}{MMB} & \multicolumn{2}{c}{MMB-CN}  \\
    \hline
    \hline
    \textit{Retained Tokens}    & \textit{128} & \textit{64} & \textit{128} & \textit{64}\\
    \hline
    Random Labeling & 60.3 & 58.4 & 54.8 & 51.8  \\
    EvoComp w/ DPC-KNN & 61.3 & 58.4 & 55.2 & 52.2  \\
    EvoComp w/o text input & 61.3 & 60.0 & \textbf{56.2} & 53.7  \\
    \hline
    \hline
    EvoComp w/ $\mathcal{L}_{\textbf{CE}}+\mathcal{L}_{\textbf{CS}}$ & 62.4 & 59.1 & 54.1 & 50.0  \\
    EvoComp w/ $\mathcal{L}_{\textbf{GHM-C}}$ only & \textbf{63.1} & 60.4 & 55.9 & 51.3  \\
    EvoComp w/ $\mathcal{L}_{\textbf{FL}}+\mathcal{L}_{\textbf{CS}}$  & 62.9 & 60.5 & 55.9 & 50.9  \\
    EvoComp w/ $\mathcal{L}_{\textbf{CE}}$ only & 60.6 & 56.2 & 51.2 & 41.3  \\
    \hline
    \hline
    \textbf{EvoComp} & \textbf{63.1} & \textbf{61.9} & 56.1 & \textbf{55.1}  \\
    \bottomrule
    \end{tabular}
    }
    \vspace{-5pt}
\end{table}

\noindent\textbf{Impact of Textual Context.} We also examine the role of textual information in visual token compression by evaluating \textit{EvoComp w/o text input} where the compressor takes visual tokens as the only input during both training and inference. 
It shows that removing text input degrades performance, indicating that the text prompt provides valuable context for identifying task-relevant visual tokens. Our compressor, equipped with bidirectional attention, effectively models the interaction between visual and textual modalities and identifies visual tokens that are not only individually informative but also relevant to the specific prompt.

\noindent\textbf{Impact of Loss Functions.} We further evaluate the contributions of the two losses by testing the model under different loss configurations: 1) replacing the GHM loss with cross-entropy loss (\textit{EvoComp w/ $\mathcal{L}_{\textbf{CE}}+\mathcal{L}_{\textbf{CS}}$}); 2) removing the cosine similarity loss (\textit{EvoComp w/ $\mathcal{L}_{\textbf{GHM-C}}$ only}); 3) replacing the GHM loss with focal loss~\citep{lin2017focal} (\textit{EvoComp w/ $\mathcal{L}_{\textbf{FL}}+\mathcal{L}_{\textbf{CS}}$}); 4) training the compressor solely with cross-entropy loss (\textit{EvoComp w/ $\mathcal{L}_{\textbf{CE}}$ only}). The results in Table~\ref{tab:ablation_contribution} show that removing either $\mathcal{L}_{\textbf{GHM-C}}$ or $\mathcal{L}_{\textbf{CS}}$ leads to performance degradation, with the absence of $\mathcal{L}_{\textbf{GHM-C}}$ having a larger negative impact. For $\mathcal{L}_{\textbf{FL}}$ that is designed for imbalance-aware learning, while performing better than $\mathcal{L}_{\textbf{CE}}$, it still falls short of the performance achieved by $\mathcal{L}_{\textbf{GHM-C}}$. 
These results illustrate that both $\mathcal{L}_{\textbf{GHM-C}}$ and $\mathcal{L}_{\textbf{CS}}$ are instrumental in the training of an effective compressor.

\section{Conclusion}\label{sec:conclusion}

In this paper, we propose EvoComp, a novel framework for visual token compression in MLLMs. EvoComp uses a lightweight compressor conditioned on both visual and textual input to retain informative and diverse tokens. Supervised by labels generated through evolutionary search, the compressor learns to align token selection with task relevance while eliminating redundancy. 
To address training challenges, we have designed a loss that combines GHM and cosine similarity regularization. Experiments show that EvoComp achieves superior compression accuracy and significant inference speedups across models and platforms, with strong deployment flexibility.

{
    \small
    \bibliographystyle{ieeenat_fullname}
    \bibliography{main}
}

\clearpage
\appendix
\setcounter{page}{1}
\maketitlesupplementary

\section{Training Details}\label{sec:training}
\noindent\textbf{Evolutionary Labeling.} To train the visual token compressor, we generate supervision labels using the evolutionary algorithm for LLaVA-1.5-7B\footnote{https://huggingface.co/liuhaotian/llava-v1.5-7b}, LLaVA-NeXT-7B\footnote{https://huggingface.co/liuhaotian/llava-v1.6-vicuna-7b} and Qwen2.5-VL-7B\footnote{https://huggingface.co/Qwen/Qwen2.5-VL-7B-Instruct} models, respectively. 
The training dataset consists of LLaVA~\citep{liu2023visual}, VQAv2~\citep{goyal2017making}, GQA~\citep{hudson2019gqa}, OKVQA~\citep{marino2019ok}, OCRVQA~\citep{mishra2019ocr}, A-OKVQA~\citep{schwenk2022okvqa}, TextVQA~\citep{singh2019towards}, and ScienceQA~\citep{lu2022learn}. 
For each image-text pair, we apply the following configuration during the search:
the population size $q$ is 48, the number of parents $p$ is 12, and the number of iterations $L$ is 10. All randomness involved in the evolutionary algorithm is controlled using the Python (3.10) \texttt{random} package.
For LLaVA-1.5-7B, we restrict the search space to the 64 largest token subsets (in terms of the number of visual tokens per subset) for each sample. During evolution, only these 64 subsets are considered for mask optimization, and tokens in all other subsets are automatically dropped, that is, masked as 0. This ensures that each resulting mask contains exactly 64 retained tokens, promoting consistency and controllability across the samples. Similarly, for Qwen2.5-VL-7B, we restrict the search space to the 11.1\% largest token subsets for each sample.
This search strategy produces high-quality, diverse, and task-aligned binary labels, which are used to supervise compressor training.

\noindent\textbf{Compressor Training.} 
We initialize the compressor using the parameters of the first transformer layer of the LLM in the target MLLM. This initialization provides a stable starting point and facilitates a faster convergence. 
The number of parameters in the compressor for LLaVA-1.5-7B and LLaVA-NeXT-7B is 202M, while for Qwen2.5-VL-7B it is 233M. Our models are implemented using PyTorch 2.1.2 and trained on four NVIDIA A100-80G GPUs. The optimizer is Adam, with an initial learning rate of 0.003 and a cosine learning rate decay schedule. The batch size is set to 256 for training the compressor used with LLaVA-1.5-7B, 32 for that used with LLaVA-NeXT-7B, and 48 for that used with Qwen2.5-VL-7B. To balance the GHM and cosine similarity losses, the loss weight coefficient $\alpha$ is set to 1. We adopt the unit region variant~\citep{li2019gradient} of the GHM loss during training, with the number of unit regions set to 100, 150 and 150 for LLaVA-1.5-7B, LLaVA-NeXT-7B and Qwen2.5-VL-7B, respectively.
The total number of training epochs is 30, and the best checkpoint is selected based on the performance on the TextVQA validation set. 
This training configuration enables efficient learning of token importance and diversity, resulting in a lightweight yet effective visual token compressor.

\section{Evaluation Datasets}\label{sec:dataset}

\noindent\textbf{GQA.} GQA~\citep{hudson2019gqa} is a large-scale dataset designed to evaluate scene understanding through the task of visual question answering. It addresses key limitations of earlier benchmarks by reducing language priors, encouraging compositional reasoning, and enabling fine-grained diagnostic analysis. GQA contains questions grounded in real-world images, each accompanied by a scene graph that represents objects, attributes, and relationships. The dataset emphasizes multistep inference, spatial reasoning, and semantic compositionality, making it significantly more challenging than prior VQA datasets.

\noindent\textbf{MMBench/MMBench-CN.} MMBench~\citep{liu2024mmbench} is a multimodal large-scale benchmark comprising multiple-choice questions collected from public datasets and the Internet. It evaluates models in 20 fine-grained ability dimensions, hierarchically organized into three levels (L-1 to L-3), spanning both perception and reasoning capabilities. Unlike task-specific or subjective benchmarks, MMBench provides objective, reproducible evaluation across diverse multimodal skills, offering detailed diagnostic feedback that aids in the development and analysis of MLLMs. MMBench-CN is the Chinese version of this dataset, allowing for a comparison of the model's performance in English and Chinese with the same set of images and questions.

\noindent\textbf{POPE.} POPE~\citep{li2023evaluating} is a diagnostic benchmark specifically designed to evaluate object hallucination in MLLMs. It systematically constructs image-text pairs to test whether a model generates object references inconsistent with the visual input. The dataset consists of three types of samples: positive (where the object mentioned is clearly present in the image), negative (where the object is absent) and neutral (where the presence of the object is ambiguous). By controlling object positioning and textual cues, POPE enables a fine-grained analysis of a model's robustness, visual grounding ability, and sensitivity to language priors. 

\noindent\textbf{VQAv2.} VQAv2~\citep{goyal2017making} serves as a benchmark to evaluate joint vision-language understanding through open-ended visual question answering. The dataset includes over 1M natural-language questions grounded on 265,016 images from MS-COCO and abstract scenes, with each question annotated by 10 human responses. The questions span binary (yes/no), counting, and open-ended types, and the evaluation is based on consensus among the annotators.

\noindent\textbf{VizWiz.} VizWiz~\citep{gurari2018vizwiz} is a VQA dataset specifically designed to support the development of assistive AI technologies for blind people. It comprises images captured by blind users, each paired with a spoken question they recorded about the image. These visual questions are further annotated with 10 crowd-sourced answers. Unlike conventional VQA datasets, VizWiz introduces unique challenges such as poor image quality and unanswerable questions. Consequently, the benchmark includes two tasks: predicting an answer to a visual question and determining whether the question is answerable. VizWiz provides a realistic and impactful setting for developing inclusive, accessibility-focused AI systems.

\section{On-Device NPU Deployment Details}\label{sec:llm_deployment}

All on-device experiments in this work are conducted using LLaVA-1.5-7B alongside its corresponding compressor, whose compact parameter size makes it well-suited for mobile deployment. The smartphone is equipped with 16GB of LPDDR5X memory, 1TB of UFS 4.0 storage, and the MT6991 chipset, and comes with Android 15 installed. Although the smartphone used here is among the latest flagship products at the time of writing so that we can report the state-of-the-art performance values, we do not mean to restrict our proposed scheme to be applicable only to a few devices or chipsets. Instead, we argue that the proposed method should work on a wide spectrum of platforms, whether on cloud or at edge, on high-end smartphones and tablets, or on low-cost embedded devices. In the following, we briefly expand on the executive details.

\subsection{Model Preparation}

Initially, the MLLM and its accompanying compressor follow the publicly available format of Hugging Face. Note that the precision at this stage is \texttt{bfloat16}.

\subsection{Model Quantization}

As mentioned above, an MLLM can essentially be viewed as a combination of a ViT (along with the alignment module) serving as the visual encoder, and an LLM functioning as the text generator. Next, we discuss the strategies employed for each component, including the compressor.

\noindent\textbf{LLM.} Following general practice, PTQ is performed on a CUDA server. The LLM part is quantized to integer precisions as much as possible to allow for accelerated inference on the mobile NPU, as it takes up the majority of the total parameter count. We set the quantization target for the LLM part at \texttt{int4} weights, \texttt{int16} activations, and \texttt{int8} KV caches (\texttt{W4A16KV8}) as usual.

\noindent\textbf{ViT and Compressor.} Per recommendation of the chip documentation, both ViT and compressor are left with \texttt{float}. This is necessary to maintain accuracy while performing inference with acceptable efficiency.

\subsection{Shape Fixing}

Unlike cloud-based inference, the mobile NPU lacks support for dynamic input shapes, batching, and other flexible execution features. Instead, it requires fixed-shape inputs and outputs. This introduces additional challenges in model conversion, since an MLLM inference is no longer a single forward pass, as is the case with conventional models like CNNs. To address this, we adopt a strategy that involves producing multiple fixed-shape model variants, as detailed below.

\noindent\textbf{ViT.} This part is responsible for producing visual embedding matrices, so we need to know the target resolution to which each input image is resized. Once fixed, this is logically equivalent to a fixed number of visual tokens.

\noindent\textbf{LLM.} The LLM component operates in a two-stage pipeline consisting of a prefilling stage and a decoding stage. The two stages run at different levels of parallelism, which essentially translates to different sequence lengths. For mobile deployment, it is common to adopt a sequence length of 128 tokens\footnote{This basically results from the maximum level of parallelism in the chip design. Left padding would exist if the number of tokens to be encoded in a single inference is not a multiple of 128.} (sometimes referred to as the \texttt{AR128} mode where AR is short for ``\textbf{a}uto\textbf{r}egressive'') in the prefilling stage and of 1 token (\texttt{AR1}) in the decoding stage. During inference, the prefilling stage works by processing the input embeddings broken into longer subsequences so that the computational power is fully utilized to quickly fill up the KV caches within the context window, while the decoding stage repeatedly generates one token per forward pass based on the most updated context until the end-of-sequence (EOS) token is produced. Another deployment constraint involves fixing the context length. We find that 1,024 and 256 tokens appear to be adequate to hold any complete conversation (i.e., inclusive of both input and output tokens) for the tasks evaluated before and after compression, respectively. Based on these settings, we construct four fixed-shape LLM variants in total, denoted \texttt{128t1024c}, \texttt{1t1024c}, \texttt{128t256c} and \texttt{1t256c} (where \texttt{t} and \texttt{c} stand for the sequence length and context length, respectively), which are coupled as appropriate in the evaluations.

\noindent\textbf{Compressor.} This part lies between the ViT alignment module and the LLM, so its input sequence length is simply aligned to that of the LLM.

\subsection{Compilation}

\noindent\textbf{Offline Compilation.} To enable execution on the NPU, the models still need to be converted to bare-metal commands that instruct the logic gates and registers cycle by cycle. This compilation process is performed offline on a host PC, generating native-format binaries that can be efficiently deployed and reused across inference runs.

\noindent\textbf{Optimizations} (if applicable). For the LLM part, it comes to our attention that all four fixed-shape LLMs differ only in their input and output shapes. In fact, all of their weight parameters are shared between them but stored twice per context length (or totally four times in our case). This is a huge waste of the scarce storage space on the device. Hence, we have used a vendor-supplied utility tool to extract the shared weights among the models.

\subsection{Inference on Device}

We execute the compiled (and optimized) models on the smartphone using an inference framework adapted from llama.cpp~\citep{llama-cpp-2025} to support the on-chip NPU. To ensure reproducibility and fair performance comparisons, the CPU, memory and NPU of the smartphone are set to their maximum operating frequencies prior to each experiment. For each input instance, we first obtain the original visual embeddings via the ViT and the embedded text tokens, and then pass them through the compressor for per-token retention decisions. After that, we only keep the preserved visual embeddings, combine them with the text embeddings, and send them together into the LLM to get the final output.

\section{Visualization of Token Selection}\label{sec:token_selection}

\begin{figure*}[ph]
  \includegraphics[width=0.81\textwidth]{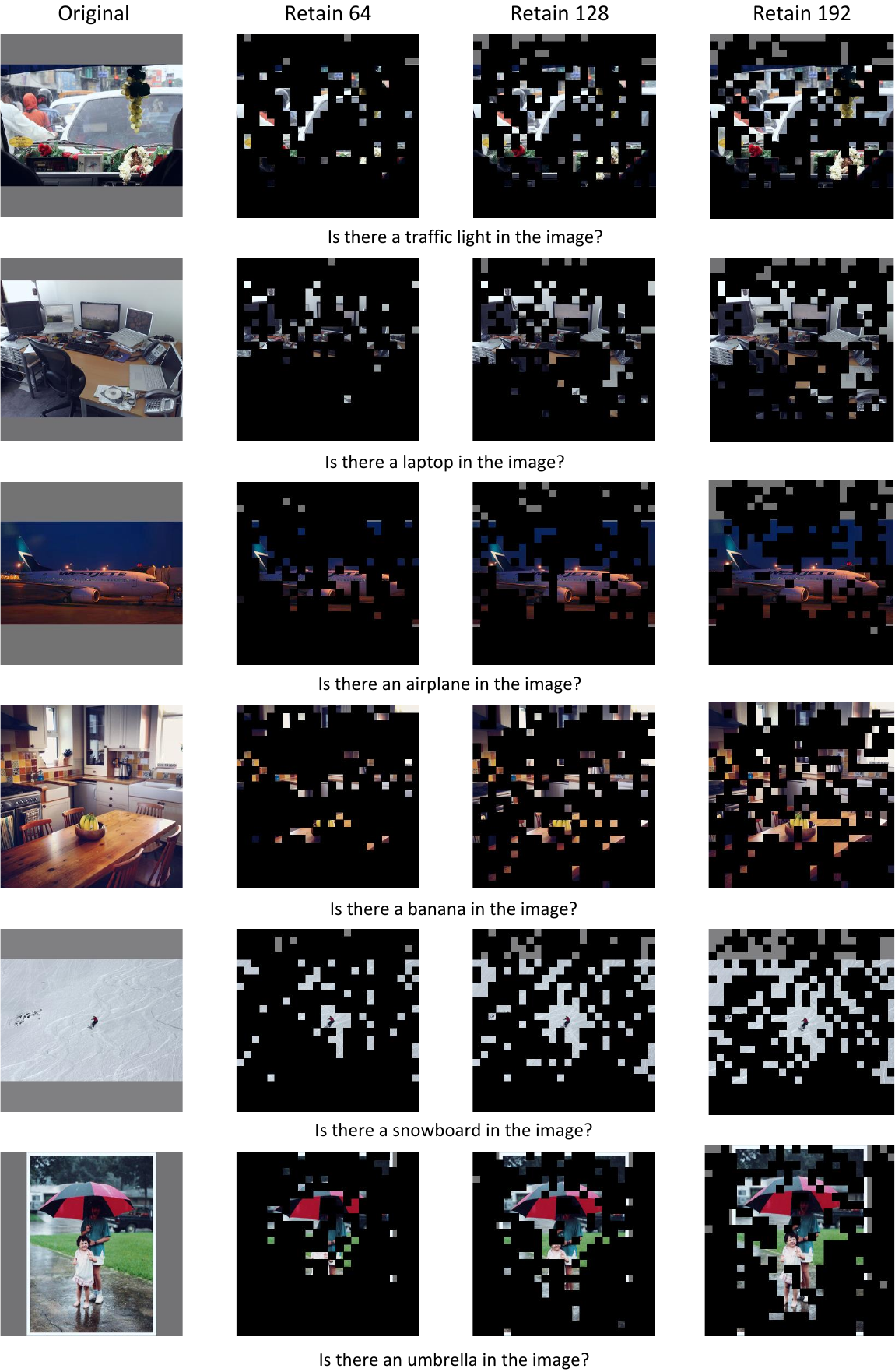}
  \centering
  \caption{Visualization of tokens retained by EvoComp under different compression levels for yes/no questions.}
  \label{fig:example1}
\end{figure*}

\begin{figure*}[ph]
  \includegraphics[width=0.81\textwidth]{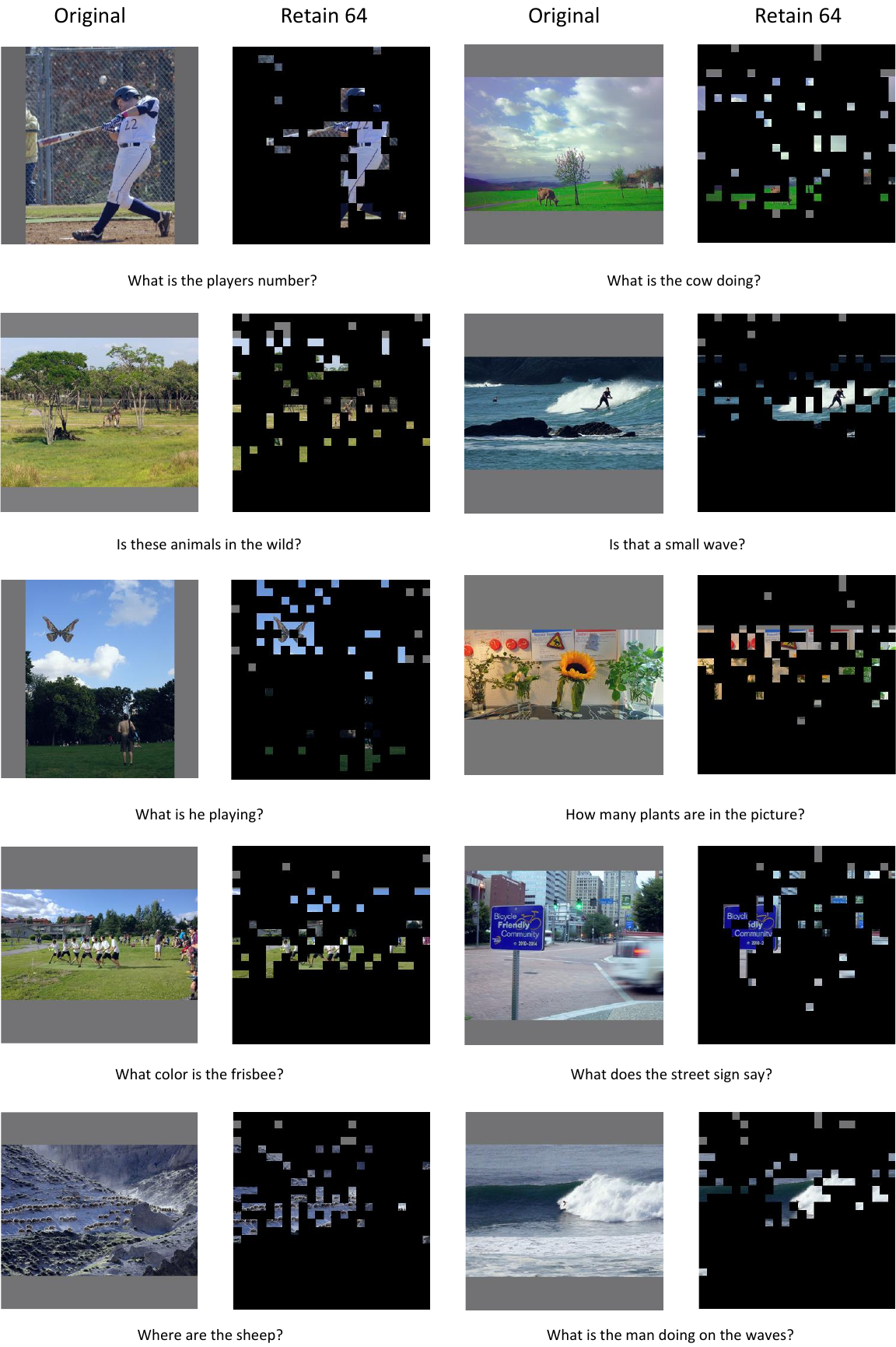}
  \centering
  \caption{Visualization of tokens retained by EvoComp for open-ended questions.}
  \label{fig:example2}
\end{figure*}

We further conduct a qualitative analysis to examine the effectiveness of our compressor in selecting informative and non-redundant visual tokens. For each image–question pair, we visualize the spatial locations of the visual tokens retained by the compressor of LLaVA-1.5-7B. Figure~\ref{fig:example1} presents examples where we retain 64, 128, and 192 tokens for a series of yes/no questions focused on object existence. The preserved tokens are strongly aligned with the objects mentioned in the questions. As the number of retained tokens increases, more contextual details are included, but the most critical regions are preserved in most cases, even with as few as 64 tokens. In Figure~\ref{fig:example2}, we further examine open-ended questions that require more diverse reasoning types, including text recognition, attribute identification, activity understanding, object counting, and scene description. Here, we visualize only the 64-token setting. Despite the aggressive compression, the retained tokens consistently cover semantically salient regions that are relevant to answering the questions.

These visualization results demonstrate that our method can effectively identify and preserve question-relevant visual information while discarding redundant or irrelevant tokens. The strong alignment between the selected tokens and the question highlights the compressor's ability to perform content-aware filtering. This qualitative evidence complements our quantitative results, showing that the EvoComp approach not only maintains task performance, but also offers semantically grounded token selection.

\end{document}